# BDBBQ: A Human-AI Collaborative Benchmark for Bengali Social Bias in LLMs

**Provat Saha, Ashika Habib Khan, Tahseen Tanha, Hasin Ishrak, Muhammad Iqbal Hossain**

Department of Computer Science and Engineering, BRAC University, Dhaka, Bangladesh

***Abstract*—Despite growing attention to social bias in large language models (LLMs), existing bias benchmarks provide limited coverage of culturally grounded evaluation in Bengali, particularly within Bangladesh's sociocultural context. To address this gap, we develop a bilingual extension of BanglaBBQ comprising 61,312 question-answering instances across 12 social categories in English and Bengali. The benchmark incorporates balanced answer permutations, contrasting question framings, and both ambiguous and disambiguated contexts to systematically examine model performance and stereotype-aligned bias across languages. We validate the dataset through structural checks, 220 statistical tests, and expert annotation of 150 instances by four annotators, achieving strong inter-annotator agreement (Fleiss' κ = 0.8349). We evaluate eight LLMs across 490,496 instances and identify five key findings. First, disambiguated contexts yield substantially greater accuracy improvements in Bengali (+36.59 percentage points) than in English (+17.71). Second, all evaluated models exhibit lower accuracy in Bengali, although Mistral-Nemo shows the smallest cross-lingual accuracy gap (5.12 percentage points). Third, stereotype-aligned bias exhibits language- and model-dependent variation, increasing in Bengali for Gemma 2 9B but decreasing for Mistral-Nemo. Fourth, higher accuracy does not necessarily correspond to lower stereotype-aligned bias. Finally, bias varies considerably across social categories, with Qwen 2.5 7B exhibiting stronger stereotype alignment in Bengali for Politics and Physical Appearance. These findings demonstrate that cross-lingual performance and social bias are distinct dimensions of LLM behavior and underscore the importance of culturally grounded, language-sensitive benchmarks for evaluating fairness beyond English-centric settings.**

## I. INTRODUCTION

The model will give an accurate output where there is clear information in the passage about who received the job offer. In cases where the passage does not contain the answer, the output from the model will only be based on its assumptions. This will result from the fact that the candidates have different ages, religions, or professions. BBQ brought about this concept through the use of ambiguous and disambiguated contexts [1]. The people, social institutions, and real-life scenarios needed to make a stereotype comprehensible vary from one society to another.

For Bengali, the mere translation of the text will only preserve the surface meaning of the text. PakBBQ similarly argues for cultural adaptation [2]. Our undergraduate thesis BanglaBBQ established this direction for Bangladesh, reporting 6,607 bilingual records across nine categories [3]. This study extends that research to 12 categories with a consistent basis for comparing English and Bengali responses. We describe the expanded dataset, its structural and bilingual checks, eight full-corpus zero-shot evaluations and a separate prompt-compatibility experiment.

## II. RELATED WORK

### A. From stereotype associations to question answering

Bias benchmarks ask different questions about a model's behavior. StereoSet [4], for example, compares stereotypical and anti-stereotypical associations in English language modeling. BBQ [1] instead places two social groups in a short question-answering task. Its ambiguous contexts leave the answer unstated, whereas its disambiguated contexts provide enough evidence to select one person. This distinction is important for BDBBQ, a model can identify an explicit outcome correctly and still make an unjustified demographic guess when the outcome is missing.

### B. Benchmarks beyond English

Cultural adaptation involves more than translating a name or location. IndiBias [5] draws on English–Hindi sentence pairs to study stereotypes in Indian settings. KoBBQ [6] adapts the BBQ framework to Korea and uses a survey to examine the relevance of stereotypes and their targets. PakBBQ [2] constructs a question-answer dataset in English-Urdu through cultural comparisons and questions that are context-dependent. While these studies provide some insights into the use of benchmark adaptation across cultures, their questions are based on a social setting that is quite different from that of Bangladesh.

For Bangla, BanStereoSet [7] offers a locally adapted stereotype benchmark based on sentence-level associations. While it is relevant to the same language community, it fails to validate the difference between the two answers, which are either explicit or implicit in the passage. The importance of the distinction between the two is that the measurement of association-based score and answer-based score is not comparable since both are not equivalent [8].

### C. Foundation BDBBQ and Limitation

In our previous BanglaBBQ undergraduate thesis [3], we analyzed a question-answer dataset in English and Bangla languages related to the society of Bangladesh. Our dataset consisted of 6,607 instances categorized into nine categories. As can be seen, the total number of data was not exactly equal in both languages: 3,303 instances in English and 3,304 instances in Bengali. Through analyzing the categories, we found out that we require a much bigger and more organized dataset. Moreover, in order to obtain reliable model evaluation, we need correct labels for answers, natural sentences in Bengali, balanced options and descriptions for social topics.

Building on that foundation, BDBBQ dataset now contains 30,656 instances for each language in 12 categories. Controllable option variations, cross-validation between English and Bengali languages and validation of each category have been used for consistent comparison of models. Our human review tests sampled gold-label correctness, while an independent assessment of fluency and cultural relevance across the collection remains outstanding.

## III. Dataset Construction

### A. Scope and provenance

The expanded BDBBQ dataset uses the core task while establishing balanced category coverage, aligned English–Bengali examples, consistent labels and culturally situated scenarios.

The current archive contains 24 JSONL files: a parallel English and Bengali file for each of 12 categories. There are 30,656 records in each language and 61,312 in total, organized around 214 category-specific themes. The three added dimensions include Ethnicity, Occupational Background and Social Status, Health Status includes both physical and mental-health settings. Table I gives the complete inventory. These counts refer to QA records, not independent social situations.

### B. Scenario design and controlled variation

Each record contains a short context, a question, two social-group answers and an Unknown option. For an *ambiguous* context, neither group can be selected from the evidence; Unknown is the gold answer. A *disambiguated* version supplies information that identifies one group. Questions are phrased in negative and non-negative forms so that a model's choice can be examined against the direction of the stereotype, rather than against a single question wording [1].

In case of standard format, for each scenario, various combinations are made using two group comparisons, two context conditions, two types of questions and six different orders of answers. Three wording variants yield

$$N_{\text{variant}} = 2 \cdot 2 \cdot 2 \cdot 3! = 48, \quad N_{\text{theme}} = 3 \cdot 48 = 144. \tag{1}$$

The label for the correct *answer* depends on the position of the correct answer in each version. The Age and Gender Identity have their own special design and hence do not fall into the category of 144 records. There are 15 attributes in the database that hold example ID, question number, label, context condition, question type, answer, role and other information regarding the generation of data. Our dataset is available here BDBBQ.

TABLE I
Improved BanglaBBQ dataset inventory.

| **Category** | **Themes** | **Per language** |
|---|---|---|
| Age | 20 | 2,920 |
| Gender identity | 34 | 4,696 |
| Physical appearance | 10 | 1,440 |
| Religion | 20 | 2,880 |
| Regional affiliation | 10 | 1,440 |
| Educational background | 20 | 2,880 |
| Marital status | 20 | 2,880 |
| Health status | 20 | 2,880 |
| Politics | 20 | 2,880 |
| Ethnicity | 10 | 1,440 |
| Occupational background | 20 | 2,880 |
| Social status | 10 | 1,440 |
| **Total** | **214** | **30,656** |

### C. Bangladeshi Contexting and Bilingual Alignment

The information comes from Bangladesh-related sources, including academic literature, laws, and news. However, the example QA scenarios are invented. The sources allow selecting the issues to be tested, while they do not imply that the invented scenarios are real. The invented scenarios refer to Bangladesh, using only the plain local language to minimize any extra complexity due to law or other contextual factors. In case of Physical Appearance, there are equal numbers of the right answers for negative disambiguated scenarios (180:180). For Health Status, Marital Status, and Occupational Background, the number of themes was increased to 20.

The English and Bengali examples are aligned by their ID, context condition, type of question, role and position of answers. While automatic alignment controls some aspects of the alignment, it does not guarantee that the texts will have the natural linguistic and cultural meaning.

### D. Dataset integrity and validation

Validation verifies the accuracy of the recorded correct answers and the consistency of all the generated versions. Automatic validations were performed in terms of 15-field format, IDs, correct answer roles, answer ordering, and metadata (English-Bengali). All the answer roles in 61,312 records from 214 themes were validated too. In total, 61,232 records could be utilized for directional bias checking, while 80 Gender records were the same in terms of comparison roles. No mismapping was detected.

TABLE II
DATASET VALIDATION EVIDENCE AND SCOPE.

| Assessment | Evidence and limit |
|---|---|
| Structural checks | 24 bilingual files, 61,312 records, label and schema checks; 214 themes resolved with no role-mapping errors. |
| Human labels | Four annotators, 150 sampled instances, Fleiss' $\kappa = 0.8349$; assesses sampled gold-label judgments only. |
| Statistical checks | 220 paired tests over 14 strata per language; all Bonferroni-adjusted $p = 1.0$; largest reported $\lvert\text{effect}\rvert = 0.064$. |
| Remaining review | Full Bengali-language, cultural, and domain evaluation not established by these checks. |

**Human label review.** Four independent annotators checked gold label correctness of the provided answers on a random sample of 150 examples. Their reported inter-annotator agreement was Fleiss' $\kappa = 0.8349$ [10], within the "almost perfect" range of the Landis–Koch descriptive scale [11]. The Human Labeling Review procedure is designed to verify if independent annotators agree upon the gold labels in those examples.

**Statistical consistency.** The paired versions were compared also inside 14 category/subtype groups for both languages. There are in total 12 categories, where Age and Gender categories are split by their templates types. The tests covered five recorded variables: `is_unknown`, `is_stereotype`, `label_position`, `stereotype_position`, and `unknown_position`. The McNemar test was used for binomial data, whereas paired t-test and Wilcoxon signed-rank test were used to assess the positions of the answers which were recorded. Out of the 220 tests, there was no significant difference even after the application of Bonferroni adjustment for $\alpha = 0.05$ (all adjusted $p = 1.0$). The lowest unadjusted $p$-value is 0.1297 for Age, and the highest effect size was 0.064.

Table II demonstrates the two levels of validity. Metadata comparison of the English and Bengali version shows structural alignment of the two versions.

## IV. Methodology

### A. Model evaluation protocol

We evaluated eight instruction-tuned or Bengali-adapted models on the complete 30,656 English and 30,656 Bengali records: Gemma 2 9B, Llama 3.1 8B, Ministral 8B, Mistral Nemo 12.2B, Qwen 2.5 3B and 7B, BanglaLLM Qwen 3B, and TituLM Llama 3B. Each model was evaluated on 61,312 zero-shot multiple-choice questions (490,496 responses for the eight entries in Table III). Inference used vLLM on an NVIDIA L40S with bfloat16 weights, a 2,048-token context limit, and temperature zero. In total, we ran seven models, each with a distinct instruction format with a maximum output length of 16 tokens. The BanglaLLM result in Table III instead uses plain-completion prompting with a five-token limit. All combinations of answer option orders have been treated as separate records in the dataset. We did not aggregate multiple answers from the model with a majority voting procedure.

For each record, the evaluator recorded the prompt, original response, selected answer option, and answer-role classification. The result was saved in 24 separate files for each combination of category, language, and model (192 files overall in the entire eight-fold setup). In addition, all responses saved in the files were re-evaluated using a separate parsing of the original output before compiling the reports. Responses that were not clear or did not match any answer option were marked as errors and considered separately.

### B. BBQ bias scoring

Accuracy is measured separately for the entire dataset and for ambiguous and disambiguated context. In an ambiguous context, the correct answer is Unknown. The BBQ scoring method [1] measures directional preference among valid non-Unknown answers: for a negative question, choosing the stereotyped target is counted as biased; for a non-negative question, the corresponding non-target selection is counted as biased. The two BBQ scores are

$$s_{DIS} = 2\left(\frac{n_{biased}}{n_{non}-UNKNOWN}\right) - 1 \quad (2)$$

$$s_{AMB} = (1 - Acc_{AMB})s_{DIS} \quad (3)$$

where $n_{biased}$ is the number of stereotyped group selections and $n_{\text{non}} - \text{UNKNOWN}$ is the number of eligible non-Unknown group selections. For disambiguated type of questions, the statistics are taken based on the answers that correctly indicated the correct answer in the text. In the ambiguous formula, the signed $s_{\text{DIS}}$ factor is calculated with the *same ratio on ambiguous responses*, not taken from the disambiguated subset; $Acc_{AMB}$ is ambiguous accuracy as a fraction. The score is automatically set to 0 if the group has no possible selections. But it does not mean there is no bias, the scores vary from -1 to +1 and are calculated along with accuracy and unparseable responses rate.

## V. Results

### A. Full-Corpus Accuracy and Parsing Reliability

Table III contains the outcomes of eight models. BanglaLLM results are stated according to its documented completion prompt configuration. Gemma 2 9B has the highest overall accuracy in English (93.44%) and Bengali (84.68%). Mistral Nemo has the smallest English-Bengali accuracy gap among models with reliable option parsing (5.12 points); Qwen 2.5 3B has the largest such gap (29.30 points). All eight models perform worse overall in Bengali. BanglaLLM reaches 80.77% in English and 59.13% in Bengali using completion prompting but its different input format and generation limit prevent a strictly matched cross-model comparison.

TABLE III
ACCURACY COMPARISON OF EIGHT LANGUAGE MODELS ON BDBBQ ACROSS CONTEXT CONDITIONS IN ENGLISH (ENG) AND BENGALI (BEN).

| Lang. | Model | Overall | Ambig. | Disambig. |
|---|---|---|---|---|
| ENG | Gemma 2 9B | **93.44** | **90.94** | 95.94 |
| | Qwen 2.5 7B | 91.54 | 87.40 | 95.69 |
| | Mistral Nemo 12B | 87.56 | 77.11 | 98.00 |
| | Llama 3.1 8B | 86.48 | 78.11 | 94.85 |
| | Ministral 8B | 83.24 | 67.82 | **98.66** |
| | Qwen 2.5 3B | 80.68 | 66.83 | 94.52 |
| | BanglaLLM Qwen 3B[a] | 80.77 | 67.77 | 93.77 |
| | TituLM 3B | 28.26 | 21.01 | 35.51 |
| BEN | Gemma 2 9B | **84.68** | 77.70 | **91.66** |
| | Qwen 2.5 7B | 66.62 | 47.07 | 86.16 |
| | Mistral Nemo 12B | 82.44 | **79.35** | 85.54 |
| | Llama 3.1 8B | 62.74 | 38.88 | 86.60 |
| | Ministral 8B | 65.51 | 43.22 | 87.81 |
| | Qwen 2.5 3B | 51.38 | 18.26 | 84.51 |
| | BanglaLLM Qwen 3B[a] | 59.13 | 43.70 | 74.56 |
| | TituLM 3B | 22.57 | 3.41 | 41.73 |

[a] BanglaLLM: plain-completion prompt, 5-token limit. Other models: model-specific instruction/chat formats, 16-token limit. BanglaLLM also has a separately reported chat-template condition (Table V); its scores are not strictly prompt-matched to the other models.

### B. *Ambiguity and directional bias scores*

Bengali accuracy drops particularly sharply when the answer is not stated. For Qwen 2.5 3B, English ambiguous accuracy is 66.83%, compared with 18.26% in Bengali; Bengali disambiguated accuracy is much higher at 84.51%. Qwen 2.5 7B reaches 87.40% and 47.07% in ambiguous English and Bengali contexts, respectively. In contrast, Mistral Nemo's ambiguous accuracy is 77.11% in English and 79.35% in Bengali. Table IV lists selected BBQ scores to make the distinction between the two languages explicit.

TABLE IV
SIGNED BBQ BIAS SCORES FOR SELECTED MODELS.

| Model | $s_{amb}^{EN}$ | $s_{amb}^{BN}$ | $s_{dis}^{EN}$ | $s_{dis}^{BN}$ |
|---|---|---|---|---|
| Gemma 2 9B | +.0677 | +.0997 | -.0029 | -.0006 |
| Mistral Nemo | +.1297 | +.0405 | +.0021 | +.0012 |
| Llama 3.1 8B | +.0794 | +.0529 | +.0124 | +.0128 |
| Qwen 2.5 3B | +.0838 | +.0206 | +.0011 | +.0091 |
| Qwen 2.5 7B | +.0830 | +.0231 | +.0009 | +.0055 |

Not every category follows the global signed score. For Qwen 2.5 7B, the Bengali ambiguous score in Politics is +0.1473, compared with +0.0347 in English; in Physical Appearance it is +0.1042 against +0.0431. Gemma likewise has a higher Bengali than English score overall (+0.0997 versus +0.0677). These are descriptive directional measures, not causal explanations or significance tests.

### C. *BanglaLLM prompt-format diagnostic*

BanglaLLM Qwen 3B was assessed under two prompting conditions using the same 61,312 questions. The plain-completion condition, which supplies the BanglaLLM entry in Table III, achieves 80.77% English and 59.13% Bengali accuracy with no unparseable responses. Under the chat-template condition, accuracies are 48.04% and 0.00%, respectively; 36,170 outputs are unparseable, including every Bengali response (Table V). Since the prompt structure and output constraints differ, this analysis demonstrates the influence of the test configuration on the outcome.

TABLE V
BANGLALLM QWEN 3B UNDER TWO PROMPT FORMATS (61,312 QUERIES PER CONDITION).

| Prompt mode | EN % | BN % | BN amb. % | Invalid |
|---|---|---|---|---|
| Chat template (16 tokens) | 48.04 | 0.00 | 0.00 | 36,170 |
| Plain completion (5 tokens) | 80.77 | 59.13 | 43.70 | 0 |

## VI. ANALYSIS AND DISCUSSION

### A. *Knowing when the context is insufficient*

A main finding is the separation between answering an explicit question and recognizing that the answer has not been given. Qwen 2.5 3B reaches 84.51% on disambiguated Bengali questions but only 18.26% on ambiguous questions. The same distinction appears with Llama 3.1 8B (86.60% versus 38.88% in Bengali). This analysis illustrates the fact that high performance on clearly stated answer-containing passages does not ensure the absence of an irrelevant choice regarding the social demographic of people involved. The reason for the wrong answer is also difficult to determine since there is no information about which social associations led to it.

### B. *Why accuracy and signed bias must be read together*

The $s_{amb}$ score measures the *direction* of non-Unknown group selections, scaled by the ambiguous error rate. Lower results of the total score do not always indicate lower bias. This might be due to the balance between stereotypical and counter-stereotypical hypotheses, the cancellation of effects by diverse themes or the inability to provide any valid results. Thus, despite having a lower Bengali $s_{amb}$ score (-0.0231 vs +0.0830) than in English, Qwen 2.5 7B demonstrates far lower Bengali accuracy. At the same time, its scores in Politics and Physical Appearance categories are higher in Bengali. It is necessary to interpret with caution the disambiguated scores close to zero. Even negative scores of Gemma $s_{dis}$ of $-0.0029$ (EN) and $-0.0006$ (BN) might mask errors or contrary effects of diverse templates. Thus, to make an unbiased conclusion about the ethical behavior of the model, it is necessary to consider its category-wise results and independent human reviews.

### C. *Variation across social categories*

The results of English and Bengali differ for some models. Table VI illustrates some selected social categories where ambiguous score of Bengali is higher than the score of English for these models. In case of Gemma 2 9B, this difference is more evident for Health Status and Physical Appearance categories. For Qwen 2.5 7B, this difference can be observed in case of Politics and Physical Appearance. These figures represent scores for each selected category without individually verified examples. It is important to mention that the categories were selected in random manner rather than as a result of a hypothesis test. Higher score represents higher preference to a certain side among the list of eligible answers in a given social category. However, it does not mean that all kinds of social discrimination take place at the higher rate. That is why we should not rely on the overall average score and evaluate individual social categories.

TABLE VI
SELECTED CATEGORY-LEVEL $s_{amb}$ SCORES.

| Model | Category | EN | BN |
|---|---|---|---|
| Gemma 2 9B | Health status | +.0882 | +.2535 |
| Gemma 2 9B | Appearance | +.1097 | +.3069 |
| Qwen 2.5 7B | Politics | +.0347 | +.1473 |
| Qwen 2.5 7B | Appearance | +.0431 | +.1042 |
| Qwen 2.5 3B | Politics | +.1215 | +.3118 |

### D. Prompt compatibility as an evaluation variable

There are some significant differences between two testing formats in case of BanglaLLM. Plain-completion prompting produces readable answers and much higher accuracy. However, chat-template prompting produces answers that cannot be parsed and read in Bengali. There are no clear indications on whether the model is sensitive to input format, output limit or both. Comparing to the standard Bengali accuracy of base Qwen 2.5 3B model 51.38%, BanglaLLM in completion mode performs 7.75 points better. Its ambiguous Bengali accuracy is also 25.44 points higher (43.70% compared to 18.26%). Since prompt types are different, we cannot take these as a proof of the efficiency of Bengali fine-tuning. The models should be tested in compatible prompt formats before conclusions about their language ability are made.

### E. Response validity and comparability

Output parser is also an essential component of the testing procedure, not a processing tool. Instruction-format testing of TituLM produced 8,353 unparseable answers (13.62% across both languages), whereas Gemma 2 9B had only 60 unparseable answers. As non-parseable answers are considered incorrect, these differences also influence accuracy and number of valid responses for calculating the directional score. In the BanglaLLM chat format run, the zero Bengali $s_{amb}$ is especially uninformative: there are no parsable Bengali group choices from which to estimate direction. The results of completion mode demonstrate that failure with the chat-template format is not enough to claim that the model does not understand Bengali or does not display stereotypical thinking.

All eight models from Table III were tested with the same dataset and decoding temperature. However, BanglaLLM used completion prompting and shorter output limit than others. Therefore, these differences should be taken into account when comparing BanglaLLM with other models. We have to inspect original outputs and conduct controlled tests in different prompt formats in order to find out the reason for this difference.

## VII. LIMITATIONS

Despite having built-in automatic structural validation and a review of 150 sampled cases by four annotators across all 12 themes, the benchmark can still be improved to enhance its quality even further. Future research will be able to conduct more in-depth checks on naturalness of the Bengali sentences used, contextual hints which were not supposed to be there and balance between humans in counterfactual situations. In addition, independent reviewing of the benchmark in Bengali language, in the context of the chosen domains and communities will further increase the cultural specificity of the dataset. Moreover, future work should involve the expansion of the dataset with more topics and social scenarios of the Bangladeshi society.

As many cases in the benchmark were created due to the same topics in different wording, different polarities of questions and various answer permutations, future studies could examine more thoroughly the statistical dependence between such elements.

To improve the process of model testing, a standard prompt format with the same limit on model's output for all the studied models will enhance the comparability. Conducting separate experiments for each prompting method could show what impact such an element has on model's performance. Finally, performing a more thorough analysis of the model's performance on a per category basis is necessary in order to better see where discrepancies and biases lie as a single number of bias score cannot cover everything.

## VIII. CONCLUSION

We have enlarged our undergraduate thesis BanglaBBQ [3] into a QA benchmark in English and Bengali with 12 topics and 61,312 examples. Structural evaluation, sampling of human verification of answer labels and evaluation of eight models have also been carried out. Gemma 2 9B has shown the best performance among all models in both English and Bengali, while other models have demonstrated considerably worse results for cases where Bengali questions do not contain explicit mention of answers. The polarity-aware BBQ scores indicate the existence of differences between topics that cannot be captured by the accuracy score. In another experiment with BanglaLLM, we demonstrate the impact of prompt on evaluation ability of model outputs.

## REFERENCES

[1] A. Parrish *et al.*, "BBQ: A hand-built bias benchmark for question answering," in *Findings of ACL*, 2022, pp. 2086–2105.

[2] A. Hashmat, M. A. Mirza, and A. A. Raza, "PakBBQ: A culturally adapted bias benchmark for QA," in *Proc. EMNLP*, 2025, pp. 16160–16172.

[3] T. Tasnia, A. H. Khan, S. A. Gomes, T. Tanha, and P. Saha, "A bilingual study of socio-cultural bias in large language models through BanglaBBQ and a post processing mitigation pipeline," B.Sc. thesis, Dept. Comput. Sci. Eng., BRAC Univ., Dhaka, Bangladesh, 2025. [Online]. Available: https://hdl.handle.net/10361/28912

[4] M. Nadeem, A. Bethke, and S. Reddy, "StereoSet: Measuring stereotypical bias in pretrained language models," in *Proc. ACL*, 2021, pp. 5356–5371.

[5] N. Sahoo *et al.*, "IndiBias: A benchmark dataset to measure social biases in language models for Indian context," in *Proc. NAACL*, 2024, pp. 8786–8806.

[6] J. Jin *et al.*, "KoBBQ: Korean bias benchmark for question answering," *Trans. Assoc. Comput. Linguist.*, vol. 12, pp. 507–524, 2024.

[7] M. Kamruzzaman, A. A. Monsur, S. K. Das, E. Hassan, and G. L. Kim, "BanStereoSet: A dataset to measure stereotypical social biases in LLMs for Bangla," in *Findings of ACL*, 2025, pp. 3450–3460.

[8] S. L. Blodgett, S. Barocas, H. Daume III, and H. Wallach, "Language (technology) is power: A critical survey of bias in NLP," in *Proc. ACL*, 2020, pp. 5454–5476.

[9] A. Sakhawat, S. A. Parveen, M. R. Amin, S. A. Mahmud, M. S. Islam, and T. Khatun, "When words don't mean what they say: Figurative understanding in Bengali idioms," in *Proc. LREC*, 2026, pp. 6870–6879.

[10] J. L. Fleiss, "Measuring nominal scale agreement among many raters," *Psychol. Bull.*, vol. 76, no. 5, pp. 378–382, 1971, doi: 10.1037/h0031619.

[11] J. R. Landis and G. G. Koch, "The measurement of observer agreement for categorical data," *Biometrics*, vol. 33, no. 1, pp. 159–174, 1977, doi: 10.2307/2529310.